\documentclass[runningheads]{llncs}

\usepackage[T1]{fontenc}

\usepackage{graphicx}
\usepackage{subcaption}
\usepackage{graphicx}
\usepackage{booktabs}
\usepackage{multirow}
\usepackage{comment}
\usepackage{amsmath,amssymb}
\usepackage{color}
\usepackage{url}
\usepackage{hyperref}
\usepackage{xcolor}
\usepackage{graphicx}
\usepackage{caption}

\definecolor{outlier}{RGB}{0,0,0}
\definecolor{car}{RGB}{255,153,0}
\definecolor{bicycle}{RGB}{0,128,192}
\definecolor{road}{RGB}{255,0,255}  
\definecolor{parking}{RGB}{255,153,204}
\definecolor{sidewalk}{RGB}{102,0,102}
\definecolor{building}{RGB}{0,204,255}
\definecolor{fence}{RGB}{51,102,255}
\definecolor{vegetation}{RGB}{0,128,0}
\definecolor{trunk}{RGB}{0,51,102}
\definecolor{terrain}{RGB}{204,255,204}
\definecolor{pole}{RGB}{204,255,255}
\definecolor{trafficsign}{RGB}{0,0,255}
\definecolor{CorrectClass}{RGB}{252,255,138}

\newif\ifreview
\reviewfalse

\ifreview
	\usepackage{lineno}

	\linenumbers
\fi

\begin{document}


\def\SubNumber{073}

\def\GCPRTrack{Special Track: Photogrammetry and remote sensing}

\title{Out-of-Distribution Detection in LiDAR Semantic Segmentation Using Epistemic Uncertainty from Hierarchical GMMs\thanks{This project is supported by the German Research Foundation (DFG), as part of the Research Training Group i.c.sens, GRK 2159, ‘Integrity and Collaboration in Dynamic Sensor Networks’.
}}

\ifreview
\else
	\titlerunning{OOD Detection in LiDAR Segmentation Using Epistemic Uncertainty}

	\author{Hanieh Shojaei Miandashti\inst{1}\orcidID{0000-0003-0968-0420} \and
	Claus Brenner\inst{1}\orcidID{0000-0002-6459-0682}}
	
	
	\institute{ Institut of Cartography and Geo-informatics, Leibniz University Hannover, 30167 Hannover, Germany
	\url{https://www.ikg.uni-hannover.de/de/} \\
	\email{\{hanieh.shojaei,claus.brenner\}@ikg.uni-hannover.de}}
\fi

\maketitle              

\begin{abstract}
In addition to accurate scene understanding through precise semantic segmentation of LiDAR point clouds, detecting out-of-distribution (OOD) objects—instances not encountered during training—is essential to prevent the incorrect assignment of unknown objects to known classes. While supervised OOD detection methods depend on auxiliary OOD datasets, unsupervised methods avoid this requirement but typically rely on predictive entropy, the entropy of the predictive distribution obtained by averaging over an ensemble or multiple posterior weight samples. However, these methods often conflate epistemic (model) and aleatoric (data) uncertainties, misclassifying ambiguous in-distribution regions as OOD. To address this issue, we present an unsupervised OOD detection approach that employs epistemic uncertainty derived from hierarchical Bayesian modeling of Gaussian Mixture Model (GMM) parameters in the feature space of a deep neural network. Without requiring auxiliary data or additional training stages, our approach outperforms existing uncertainty-based methods on the SemanticKITTI dataset—achieving an 18\% improvement in AUROC, 22\% increase in AUPRC, and 36\% reduction in FPR95 (from 76\% to 40\%), compared to the predictive entropy approach used in prior works.

\keywords{Out-of-distribution detection  \and Semantic segmentation \and Uncertainty estimation  \and LiDAR scene understanding.}
\end{abstract}

\section{Introduction}
\label{sec:introduction}
Recent advances in deep learning have substantially improved the accuracy of semantic segmentation for LiDAR point clouds, along with more reliable methods for estimating prediction uncertainty. These developments are critical for safety-critical applications such as autonomous driving, where accurate perception and awareness of uncertainty help mitigate decision-making risks. However, real-world environments often include OOD instances—inputs that differ significantly from the in-distribution (ID) data seen during training. Deep models tend to make overly confident yet incorrect predictions on such OOD inputs, misclassifying them as known classes. While uncertainty estimation captures the likelihood of error in predictions, OOD detection aims to determine whether an input belongs to the training distribution. Robust OOD detection is thus essential for identifying and excluding anomalous inputs, ultimately improving the safety and reliability of autonomous systems.

Existing research on OOD detection can be broadly categorized into two approaches: supervised methods that rely on auxiliary data and unsupervised methods that operate without requiring additional data. The limitations of supervised methods are discussed in Section~\ref{sec:outlier exposure}, including their dependence on auxiliary datasets, which may be unavailable in certain applications. In contrast, our work focuses on unsupervised methods that identify OOD samples by estimating uncertainty, based on the intuition that highly uncertain predictions are indicative of OOD inputs. Recent Bayesian approaches, based on the entropy of the averaged decision (predictive entropy) from Monte Carlo dropout sampling \cite{gal2015dropout} or deep ensembles \cite{lakshminarayanan2017simple}, have been considered to compute uncertainty, which can be later used to detect OOD samples.

It has been argued by \cite{kirsch2021pitfalls} that predictive entropy is an unreliable metric for OOD detection because it conflates two distinct types of uncertainty: epistemic uncertainty, caused by limited knowledge of model parameters,
and aleatoric uncertainty, stemming from inherent data noise. While OOD samples are characterized by high epistemic uncertainty, ambiguous ID samples exhibit high aleatoric uncertainty, causing both cases to potentially yield high predictive entropy. This overlap reduces the effectiveness of predictive entropy for distinguishing OOD from ambiguous ID inputs. Consequently, predictive entropy is only a reliable OOD indicator when aleatoric uncertainty is negligible, an assumption rarely valid in practice.

To address this limitation, \cite{kirsch2021pitfalls,smith2018understanding} have proposed explicitly isolating epistemic uncertainty from the total predictive uncertainty. A principled approach is to use the mutual information (MI) between the model parameters \( \theta \) and the predicted label \( y \), conditioned on the input \( x \), as a measure of epistemic uncertainty:
\[
\underbrace{\mathbb{I}[y; \theta \mid x]}_{\text{Epistemic Uncertainty (MI)}} = 
\underbrace{\mathcal{H}(p(y \mid x))}_{\text{Predictive Entropy}} - 
\underbrace{\mathbb{E}_{\theta} \left[ \mathcal{H}(p(y \mid x, \theta)) \right]}_{\text{Aleatoric Uncertainty}}.
\]

\noindent
Here, the first term represents the entropy of the marginal predictive distribution and captures the total uncertainty (predictive entropy) associated with the prediction. The predictive distribution for a given input \( x \) is obtained by marginalizing over the posterior distribution of the model parameters \( p(\theta \mid \mathcal{D}) \) where \( \mathcal{D} \) denotes the training dataset, i.e., \( p(y \mid x, \mathcal{D}) = \mathbb{E}_{\theta \sim p(\theta \mid \mathcal{D})} [p(y \mid x, \theta)] \).
The second term quantifies the aleatoric uncertainty, the expected data uncertainty under fixed model parameters. This type of uncertainty is only defined for ID inputs, as it relies on class ambiguity captured within the training data distribution. For OOD inputs, where \( p(x) \approx 0 \), the conditional distribution \( p(y \mid x) = \frac{p(x, y)}{p(x)} \) becomes ill-defined, rendering both aleatoric uncertainty and mutual information undefined. Therefore, in practical settings using deep ensembles or Monte Carlo Dropout (MC Dropout), predictive entropy remains the only accessible uncertainty estimate for OOD detection, although it must be interpreted cautiously, as it reflects both epistemic and aleatoric components. As such, predictive entropy serves only as an upper bound on epistemic uncertainty.

With this work, we focus on an unsupervised pixel-wise OOD detection based on the epistemic uncertainty derived from the hierarchical Bayesian uncertainty measurement from the GMM feature space of a deep model. Unlike purely discriminative models that directly learn decision boundaries between classes \( p(y \mid \mathbf{x}) \), \cite{shojaei2025hierarchical} proposed employing GMMSeg \cite{liang2022gmmseg}, a generative classifier, to model class-specific Gaussian distributions in the feature space and model the uncertainty over the parameters of this GMM (mean and covariance). As a result, they propose epistemic uncertainty as the variability of the classification decisions made by the GMM samples in the feature space. This allows us to move beyond predictive entropy, which entangles aleatoric and epistemic components, and apply epistemic uncertainty for OOD detection. 

It is noteworthy that, although this paper focuses on LiDAR data, the data is represented in the form of range-view images; thus, we refer to image elements as pixels rather than points. These pixels can subsequently be projected back into 3D space to reconstruct the original point cloud. 

To summarize, we make two main claims: First, we demonstrate that epistemic uncertainty provides a sharper distinction between ID and OOD samples compared to predictive entropy, leading to improved OOD detection performance. Second, our method is fully unsupervised—it does not require auxiliary OOD data and avoids retraining the segmentation network, making it practical and efficient for real-world deployment.

\section{Related Works}
OOD detection, also known as novelty or anomaly detection, has been extensively studied in the context of classification tasks, which are image-level OOD detection \cite{hendrycks2019using,lee2018simple,rezende2015variational,shojaei2024uncertainty}. However, in this section, we focus on OOD detection methods specifically developed for semantic segmentation, where the domain is pixel-level, rather than image-level OOD detection.

\subsection{OOD Detection using Outlier Exposure}
\label{sec:outlier exposure}
Supervised OOD detection methods train models to distinguish ID from OOD samples using auxiliary datasets, a strategy known as outlier exposure \cite{hendrycks2018deep}. Some adopt a separate discriminative model \cite{bevandic2018discriminative}, while others integrate OOD detection into the segmentation network via multi-task learning \cite{bevandic2019simultaneous}. Williams et al. \cite{williams2021fool} further extend this by combining contrastive loss with data augmentation for joint segmentation and OOD detection. Hornaue and Belagiannis \cite{hornauer2023heatmap} attach a decoder to a fixed backbone to generate heatmaps with low responses for ID inputs and high activations for OOD inputs.

Despite their advantages, supervised methods have notable limitations: they (1)~depend on auxiliary data, which is often unavailable or difficult to obtain, particularly in LiDAR-based segmentation \cite{huang2022out}; (2)~make assumptions about the nature of potential anomalies that might not align with real-world cases \cite{vojivr2024pixood}; (3)~require retraining segmentation models as multi-task architectures, potentially degrading performance \cite{vandenhende2021multi}; (4)~their reliance on specific OOD training samples limits their ability to generalize to unseen or diverse types of OOD data, since even diverse auxiliary outliers during training may fail to cover the full range of real-world OOD instances\cite{di2021pixel}.

\subsection{OOD Detection using Uncertainty Estimation}
Assuming that OOD inputs should exhibit higher uncertainty (i.e., lower confidence) than ID data, early works employed softmax-based scores for OOD detection, with Maximum Softmax Probability (MSP) proposed as a baseline method \cite{hendrycks2016baseline}. However, softmax outputs not only conflate epistemic and aleatoric uncertainty \cite{kirsch2021pitfalls}, but also suffer from poor calibration, often tend to be overconfident \cite{guo2017calibration,jiang2018trust,minderer2021revisiting}, even for inputs far from the training distribution \cite{nguyen2015deep}, which limits their effectiveness in distinguishing ID from OOD samples. To overcome these limitations, ODIN \cite{liang2017enhancing} introduced temperature scaling for confidence calibration and input perturbations to increase the separability between ID and OOD samples. Although ODIN improves detection performance without requiring model retraining, it does depend on access to OOD data for parameter tuning. 

Bayesian methods offer a principled approach to uncertainty estimation by placing a prior over model parameters and marginalizing predictions over the resulting posterior. As exact inference is intractable in deep networks, approximations such as MC Dropout\cite{gal2015dropout}, which performs multiple forward passes with dropout enabled at inference time, and deep ensembles \cite{lakshminarayanan2017simple} consist of independently trained networks with different random initializations, are commonly used. However, in addition to the limitations of predictive entropy derived from these methods, as discussed in Sec.~\ref{sec:introduction}, deep ensembles also require multiple independent training runs, significantly increasing computational cost.

Our work belongs to the category of unsupervised OOD detection methods that do not require additional data, and it also addresses key limitations of deep ensembles and MC Dropout, specifically, the need for retraining to produce uncertainty estimation and reliance on predictive entropy for OOD detection.

\section{Methodology}
\begin{figure*}
    \centering
    \includegraphics[width=0.99\linewidth]{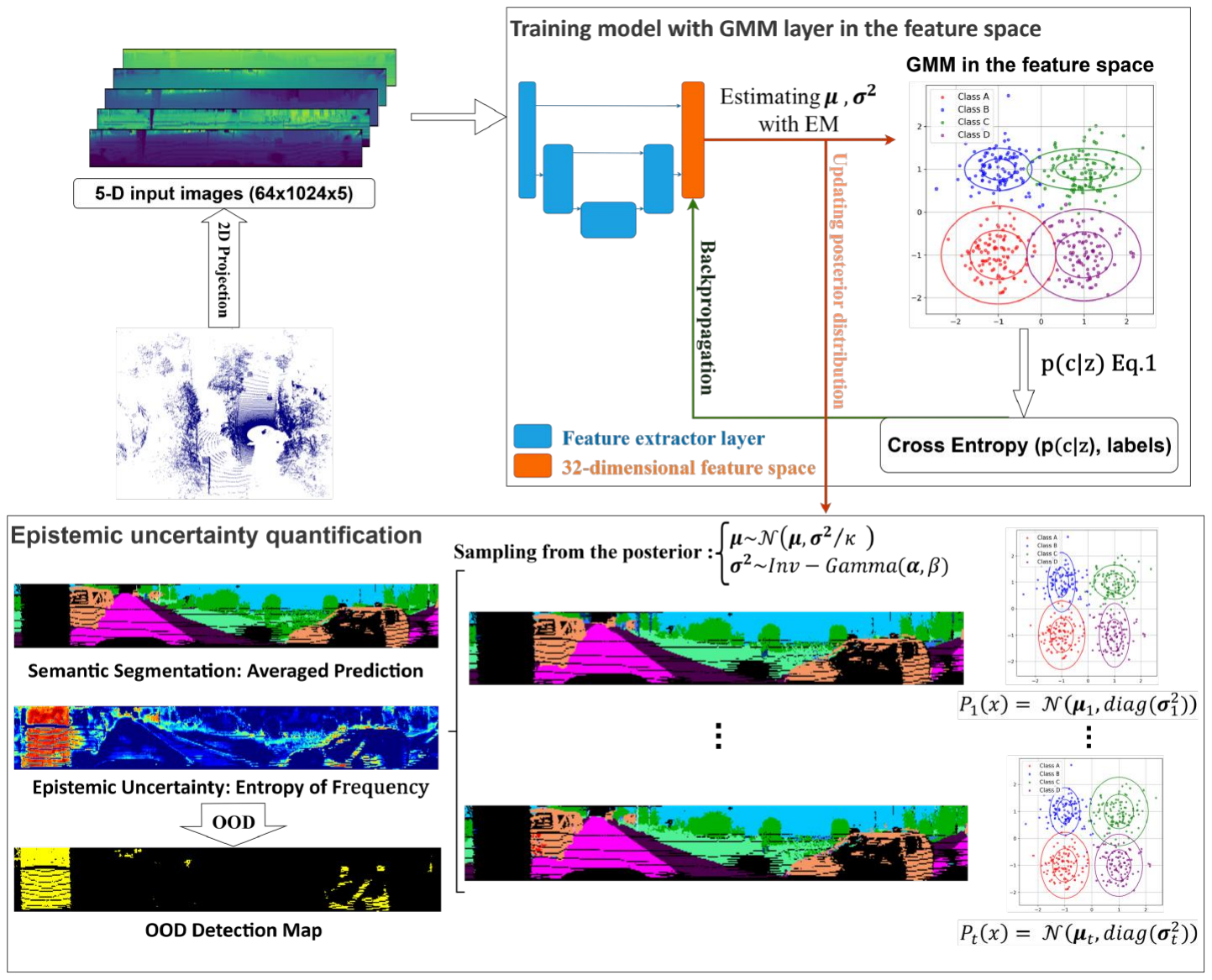}
    \caption{Proposed epistemic uncertainty-based OOD detection. A deep neural network extracts a 32-dimensional feature space from 5D LiDAR input. Class-conditional GMMs model the feature distribution per class. During inference, epistemic uncertainty is quantified by aggregating multiple predictions obtained through sampling GMM parameters (mean and variance) from their posterior distributions. Pixels exhibiting higher epistemic uncertainty scores are considered more likely to represent OOD instances.}
    \label{fig:flowchart}
\end{figure*}

To advance beyond conventional discriminative approaches for LiDAR point cloud segmentation, we adopt a GMM semantic segmentation approach inspired by GMMSeg \cite{liang2022gmmseg}. Unlike some methods that naively assume that features follow Gaussian distributions in the feature space \cite{shojaei2024uncertainty}, GMMSeg explicitly models each semantic class with a dedicated Gaussian distribution. As illustrated in the methodology framework (Fig.~\ref{fig:flowchart}), our model encodes raw LiDAR point clouds into structured multi-channel image representations, which are then processed through a deep feature extractor to obtain a compact multi-dimensional feature space. Within this feature space, the joint distribution \( p(\mathbf{z}, c) \) of pixel-wise features \( \mathbf{z} \) and semantic class labels \( c \) is modeled using class-conditional Gaussian Mixture Models (GMMs). This generative component captures the underlying feature distributions for each semantic class, while a discriminative training objective employs cross-entropy loss on the posterior class probabilities \( p(c \mid \mathbf{z}) \) derived from the GMM responsibilities, ensuring the learned features are both semantically meaningful and well-structured. Building on the epistemic uncertainty modeling framework from \cite{shojaei2025hierarchical}, we further propose an unsupervised OOD detection employing epistemic uncertainty estimates derived from the GMM parameters, enabling robust identification of OOD inputs in LiDAR semantic segmentation.

\subsection{Generative-Discriminative Framework Semantic Segmentation}
Given the LiDAR point cloud projected into a range-view image and building on GMMSeg \cite{liang2022gmmseg}, we employ a deep neural network to extract pixel-wise feature representations $\mathbf{z} \in \mathbb{R}^D$. We model the pixel-wise data distribution (joint distribution) for each class \( c \) $\in \{1, \dots, C\}$ in a \( D \)-dimensional feature space using a weighted mixture of \( M \) multivariate Gaussian components in each class as: 

\[
p(\mathbf{z} \mid c) = \sum_{k=1}^{K} \pi_{k}^{(c)} \mathcal{N}(\mathbf{z} \mid \boldsymbol{\mu}_{k}^{(c)}, \boldsymbol{\Sigma}_{k}^{(c)}),
\]

where $\pi_{k}^{(c)}$, $\boldsymbol{\mu}_{k}^{(c)}$, and $\boldsymbol{\Sigma}_{k}^{(c)}$ denote the mixture weight, mean, and covariance of the $k$-th Gaussian component for class $c$, respectively. These parameters are estimated via the Expectation-Maximization (EM) algorithm during training. As \cite{liang2022gmmseg} considered a uniform prior on the class mixture weights, i.e., \( p(c) = \frac{1}{C} \), allows us to simplify the class posterior as:

\begin{equation}
\label{eq:resp}
p(c \mid \mathbf{z}) = \frac{p(\mathbf{z} \mid c)}{\sum_{c'} p(\mathbf{z} \mid c')}.
\end{equation}

Simultaneously, the feature extractor is trained with the discriminative (cross-entropy) loss, i.e., maximizing the conditional likelihood $p(c \mid \mathbf{z})$, also referred to as responsibility value derived from GMM, promoting class separability in the learned feature space. 

\subsection{Bayesian Modeling of GMM Parameters}
To capture epistemic uncertainty, following the approach in \cite{shojaei2025hierarchical}, we adopt a hierarchical Bayesian uncertainty modeling by placing conjugate priors over the GMM parameters to model their uncertainty and updating them in a Bayesian manner to approximate the posterior distribution over the GMM parameters. 

Following \cite{liang2022gmmseg}, each multivariate Gaussian distribution is modeled as \( D \) independent univariate Gaussians, where \( D \) is the dimensionality of the feature space. During training, the mean and variance of each Gaussian component are estimated from the training data. However, unlike standard GMMs that rely on point estimates, the Bayesian approach models these parameters as random variables. Specifically, the mean and variance of each component are treated as random variables, modeled by a Gaussian and an Inverse Gamma distribution, respectively.

To enable Bayesian inference, a Gaussian–Inverse Gamma prior is placed over these parameters, serving as a conjugate prior for the unknown Gaussian mean and variance. This formulation allows the model to compute a posterior distribution over the mean and variance for each component. At inference time, the mean $\boldsymbol{\mu}_{k}^{(c)}$ and variance $\boldsymbol{\sigma^2}_{k}^{(c)}$ of each component are treated as random variables drawn from the learned posterior.

\[
\boldsymbol{\mu}_{k}^{(c)} \sim \mathcal{N}(\boldsymbol{\mu}_0, \boldsymbol{\sigma^2}_{k}^{(c)} / \kappa_0), \quad \boldsymbol{\sigma^2}_{k}^{(c)} \sim \text{Inv-Gamma}(\boldsymbol{\alpha_0}, \beta_0),
\]

\noindent where $\boldsymbol{\mu}_0$, $\kappa_0$, $\boldsymbol{\alpha_0}$, and $\beta_0$ are hyperparameters of the prior distributions. During training, these priors are updated to posteriors based on the observed data as explained in \cite{shojaei2025hierarchical}, allowing for modeling the distribution over the parameters of the GMM.

\subsection{Inference and Uncertainty Estimation}
At inference time, we sample multiple sets of GMM parameters \( (\boldsymbol{\mu}_{k}^{(c)}, \boldsymbol{\sigma^2}_{k}^{(c)}) \) from their respective posterior distributions. For each sample, we compute the class-conditional density \( p(\mathbf{z} \mid c) \) for a given feature vector \( \mathbf{z} \in \mathbb{R}^d \). 
Given an ensemble of \( n \) GMMs, the predicted class for each sample is selected as the one with the highest density. The final prediction for each pixel is obtained by majority voting across the ensemble outputs.

To quantify uncertainty, we estimate the class-wise frequency distribution of the predicted labels \( y \) over all samples. We then compute the entropy of this empirical distribution (\( \bar{p}_c = \frac{1}{n} \sum_{i=1}^{n} \mathbb{I}[y^{(i)} = c] \)) as:
\[
\mathcal{H}[y \mid \mathbf{z}] = -\sum_{c=1}^{C} \bar{p}_c \log \bar{p}_c,
\]
where \( \bar{p}_c \) denotes the relative frequency (i.e., empirical probability) of class \( c \) among the sampled predictions. This entropy reflects the degree of disagreement across sampled GMMs, capturing the epistemic uncertainty induced by variation in the GMM parameters. Higher entropy values indicate greater model uncertainty, which typically correlates with OOD regions.

\subsection{OOD Detection via Epistemic Uncertainty}
We use the computed epistemic uncertainty to detect OOD regions in the input data, based on the assumption that higher epistemic uncertainty indicates a greater likelihood of the sample being OOD. Pixels with entropy values exceeding a predefined percentile-based threshold are classified as OOD. Specifically, the threshold is set such that the top 5\% of pixels with the highest epistemic uncertainty are labeled as OOD samples. Nonetheless, to ensure robustness, we also assess epistemic uncertainty using threshold-independent evaluation metrics, as detailed in the following section (\ref{sec:evaluation metrics}).

\section{Experiments}
\paragraph{Experimental setup and dataset}
We consider the GMM feature space using the 32-dimensional penultimate layer of a convolutional neural network, SalsaNext \cite{cortinhal2020salsanext}, and conduct experiments on the SemanticKITTI benchmark dataset \cite{behley2019semantickitti}, which provides 3D LiDAR semantic segmentation annotations for 19 semantic classes along with an outlier class which is excluded during training. For OOD detection, we use the outlier class, which includes diverse objects not belonging to the 19 standard classes, such as trash bins, placards, animals, and other uncommon street elements.

When projecting 3D point clouds into 2D range-view images, due to occlusion and limited field-of-view, some pixels correspond to no 3D points and are empty. It is noteworthy that these pixels are ignored during training and excluded from OOD evaluation to avoid confusion with true outliers.

Input images are represented as tensors of shape $64 \times 1024 \times 5$, where the five channels correspond to the $x, y, z$ coordinates, intensity, and range. Forwarding into the network, the extracted features have a shape of $64 \times 1024 \times 32$, corresponding to the $32$-dimensional feature space. For each semantic class, we model the distribution of features using a GMM with two components (a hyperparameter that can be tuned). As a result, the mean and variance parameters for each class are of shape $32 \times 2$, representing two Gaussian components across the $32$ feature dimensions. Once the posterior distribution of Gaussian-Inverse Gamma is computed in the last training epoch, which allows the posterior to see the whole data once, we derive the entropy of the frequency of 20 sampled GMMs. 

We employ epistemic uncertainty for OOD detection by framing it as a binary classification task, where pixels with the top 5\% highest uncertainty values are identified as OOD. While access to OOD annotations allows for the determination of an optimal threshold, such labels are typically unavailable in real-world scenarios. Therefore, to simulate a more realistic deployment setting, we evaluate our method using an unsupervised percentile-based thresholding strategy, independent of ground-truth for OOD annotations during inference. It is important to note that this thresholding method is used solely for generating the OOD maps (Fig.~\ref{fig:OOD_comparison}), while quantitative evaluation is conducted using threshold-independent metrics (Table~\ref{tab:ood_table}).

\paragraph{Evaluation Metrtics}
\label{sec:evaluation metrics}
To evaluate OOD detection performance, we report standard metrics that are threshold-independent, including AUROC, AUPRC, and FPR95. While AUROC is widely used, AUPRC (also referred to as average precision) is more informative in the presence of class imbalance, as it focuses on precision and recall for OOD pixels \cite{chawla2004special}. Additionally, we report FPR95, which quantifies the false positive rate when the true positive rate for OOD pixels reaches 95\%, offering a safety-relevant measure of detection reliability. Moreover, the accuracy of semantic segmentation is evaluated using the mean Intersection over Union (mIoU) metric.

\paragraph{Baselines}
We compare our approach with existing OOD detection methods based on uncertainty estimation that do not rely on auxiliary data. Specifically, we consider Maximum Softmax Probability (MSP) \cite{hendrycks2016baseline} and ODIN \cite{liang2017enhancing} as representative single deterministic methods, and MC Dropout \cite{gal2015dropout} and deep ensembles (DE) \cite{lakshminarayanan2017simple} as representative Bayesian neural network (BNN) approaches. Additionally, we include the original GMMSeg \cite{liang2022gmmseg} baseline without the proposed hierarchical epistemic uncertainty estimation. For a fair comparison, all methods use SalsaNext \cite{cortinhal2020salsanext} as the backbone architecture.

\subsection{Quantitative Results}
Evaluating the performance of our proposed method against established OOD detection baselines of MSP, ODIN, GMMSeg, MC Dropout, and DE, as shown in Table~\ref{tab:ood_table}, highlights that generally our approach consistently outperforms all baselines across all OOD detection metrics.

Specifically, our method achieves the highest AUROC of 91.06\%, substantially surpassing GMMSeg (87.62\%) and DE (73.03\%). In terms of AUPRC, our method reaches 37.67\%, improving over the next best (GMMSeg, 26.14\%) by over 11 percentage points. Moreover, our method attains the lowest FPR95 of 40.14\%, reflecting its effectiveness in reducing false positives among in-distribution pixels.

In addition to OOD detection performance, our method also yields the highest mIoU (57.71\%) among all compared methods. This slight improvement in segmentation accuracy is attributed to the voting-based prediction strategy introduced through our uncertainty-aware framework.

\begin{figure*}[ht]
    \centering
    \begin{subfigure}[b]{0.49\linewidth}
        \centering
        \includegraphics[width=\linewidth]{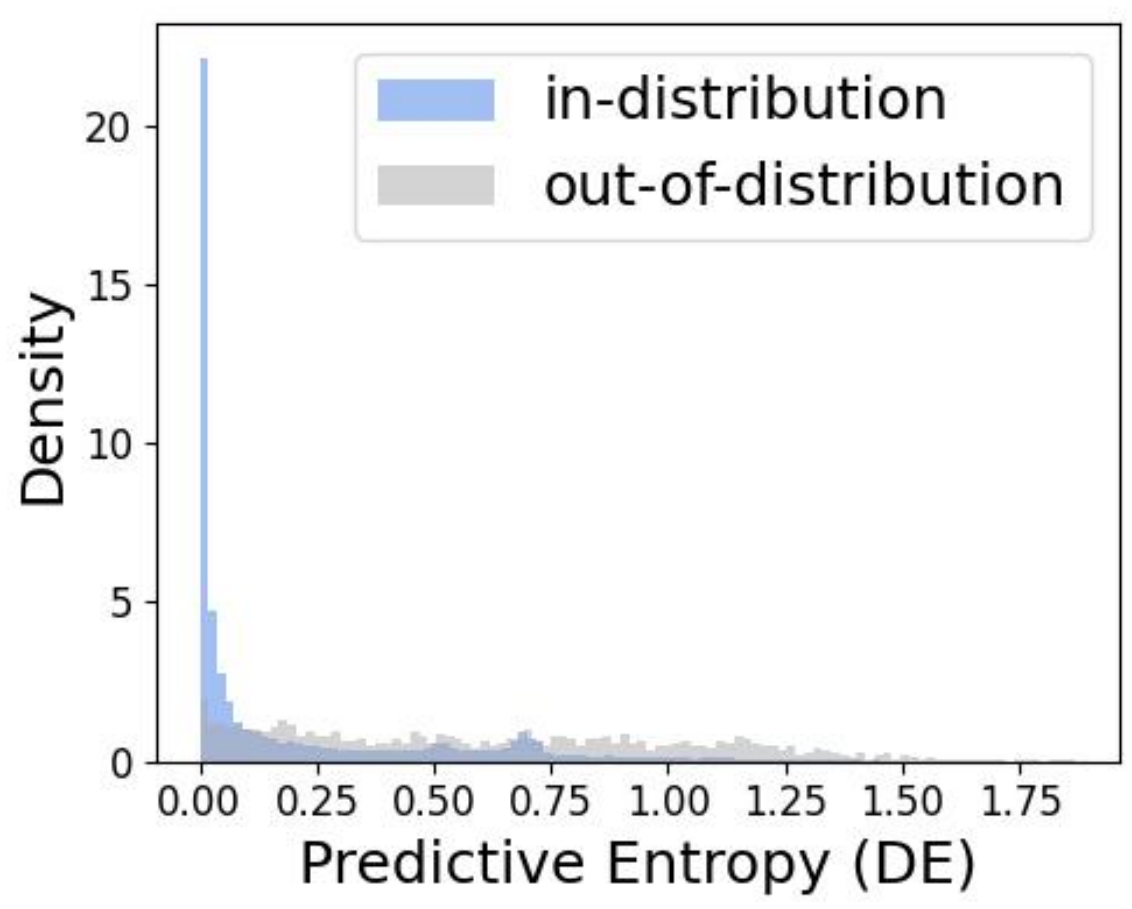}
        \caption{Predictive entropy (DE)}
        \label{fig:entropy_histograms-a}
    \end{subfigure}
    \begin{subfigure}[b]{0.49\linewidth}
        \centering
        \includegraphics[width=\linewidth]{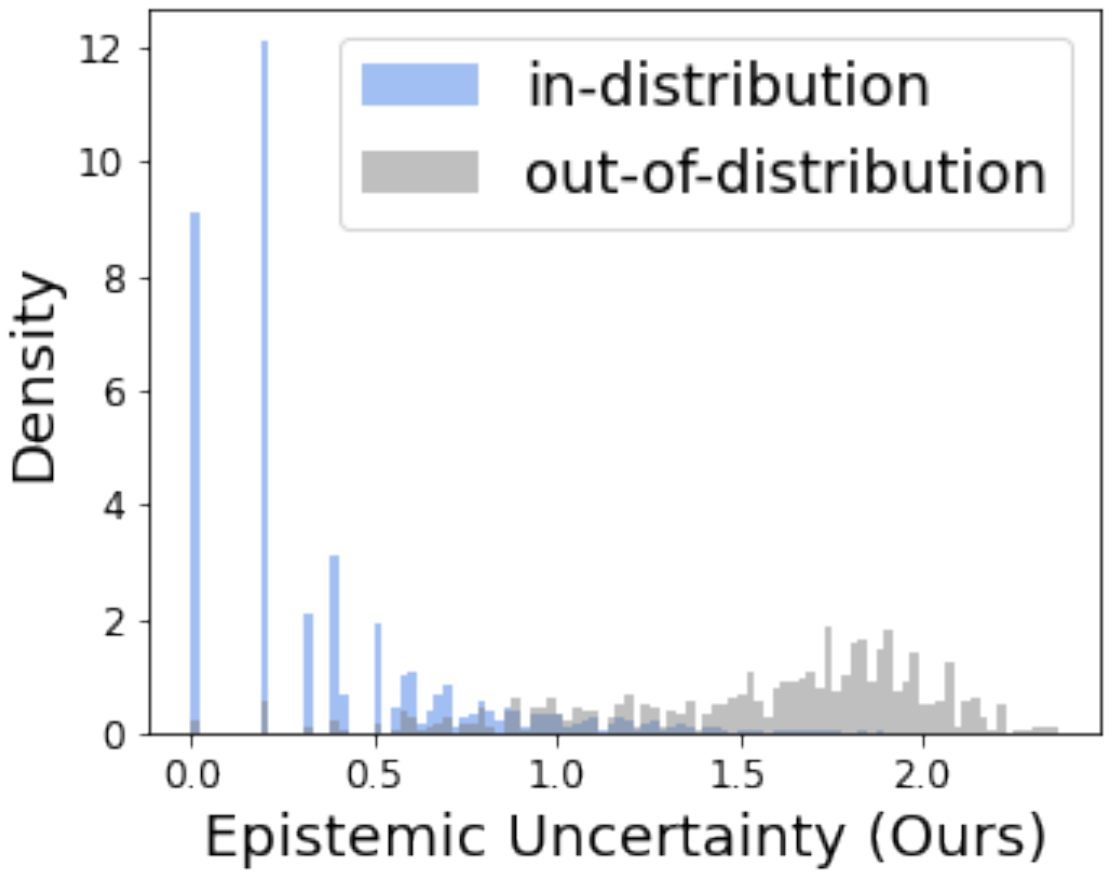}
        \caption{Epistemic uncertainty (Ours)}
        \label{fig:entropy_histograms-b}
    \end{subfigure}
    \caption{Comparison of predictive entropy and epistemic uncertainty for OOD detection. The proposed epistemic uncertainty measure offers a more precise separation between ID and OOD samples.}
    \label{fig:entropy_histograms}
\end{figure*}

To gain further insight into how epistemic uncertainty distinguishes OOD from ID samples, we compare the distributions of predictive entropy and epistemic uncertainty in Figure~\ref{fig:entropy_histograms}, derived from DE and our proposed approach, respectively. While both methods show generally higher uncertainty values for OOD samples shown in gray color, Figure~\ref{fig:entropy_histograms-a} reveals that predictive entropy from DE produces a substantial overlap between ID and OOD distributions, particularly in the mid-to-high entropy range. In contrast, Figure~\ref{fig:entropy_histograms-b} demonstrates that our epistemic uncertainty estimates produce a clearer separation, with ID samples concentrated around low uncertainty values and OOD samples distributed more distinctly across higher values. This improved separation highlights the advantage of modeling epistemic uncertainty directly via hierarchical Bayesian inference in the GMM feature space, enabling more reliable OOD detection.

\renewcommand{\arraystretch}{1.5}
\begin{table}[t]
\centering
\caption{Comparison of OOD detection methods. Our approach achieves higher AUROC and AUPRC (AP), along with lower FPR95 compared to existing methods. Additionally, mIoU shows a slight improvement due to the voting-based prediction strategy in our approach. All metrics are represented in percentages.}
\label{tab:ood_table}
\resizebox{0.90\linewidth}{!}{%
\begin{tabular}{llcccc}
\toprule
\textbf{Method} & \textbf{Uncertainty Type} & 
\textbf{↑AUROC} & \textbf{↑AUPRC(AP)} & \textbf{↓FPR95} & \textbf{↑mIoU} \\
\midrule
MSP & Deterministic Entropy & 70.41 & 10.90 & 76.00 & 56.37 \\
ODIN & Deterministic Entropy & 73.74 & 12.45 & 75.54 & 56.37 \\
MCDropout & Predictive Entropy & 73.64 & 13.65 & 75.92 & 57.15 \\
DE & Predictive Entropy & 73.03 & 16.14 & 76.48 & 57.17 \\
GMMSeg & Deterministic Entropy & 87.62 & 26.14 & 48.84 & 57.60 \\
Ours & Epistemic Uncertainty & \textbf{91.06} & \textbf{37.67} & \textbf{40.14} & \textbf{57.71} \\

\bottomrule
\end{tabular}
}
\end{table}

\subsection{Qualitative Results}
\begin{figure}
    \centering
    \begin{subfigure}{0.45\textwidth}
        \centering
        \includegraphics[width=\textwidth]{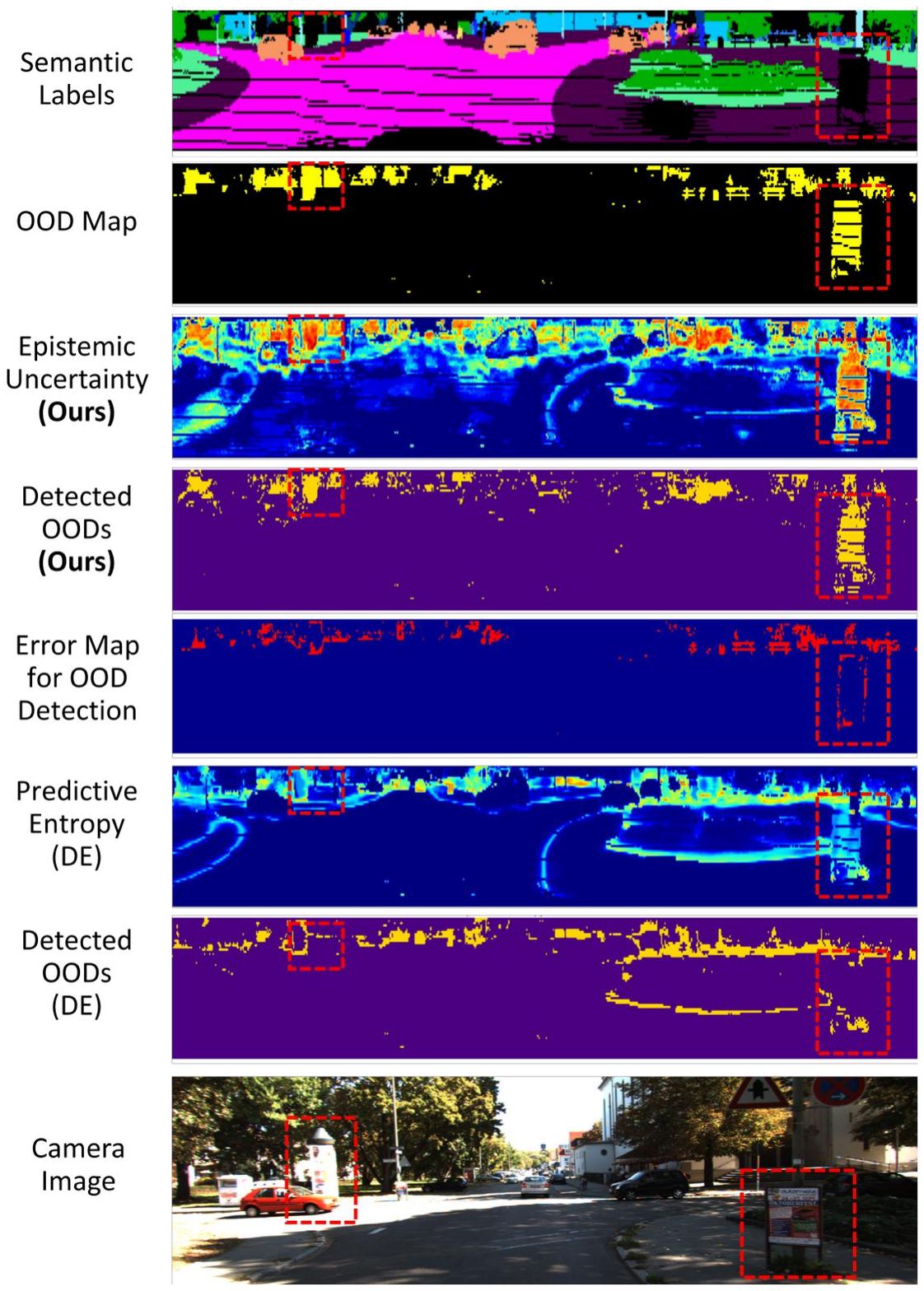}
        \caption{OOD object: Placards}
        \label{fig:OOD_comparison-a}
    \end{subfigure}
    \hfill
    \begin{subfigure}{0.45\textwidth}
        \centering
        \includegraphics[width=\textwidth]{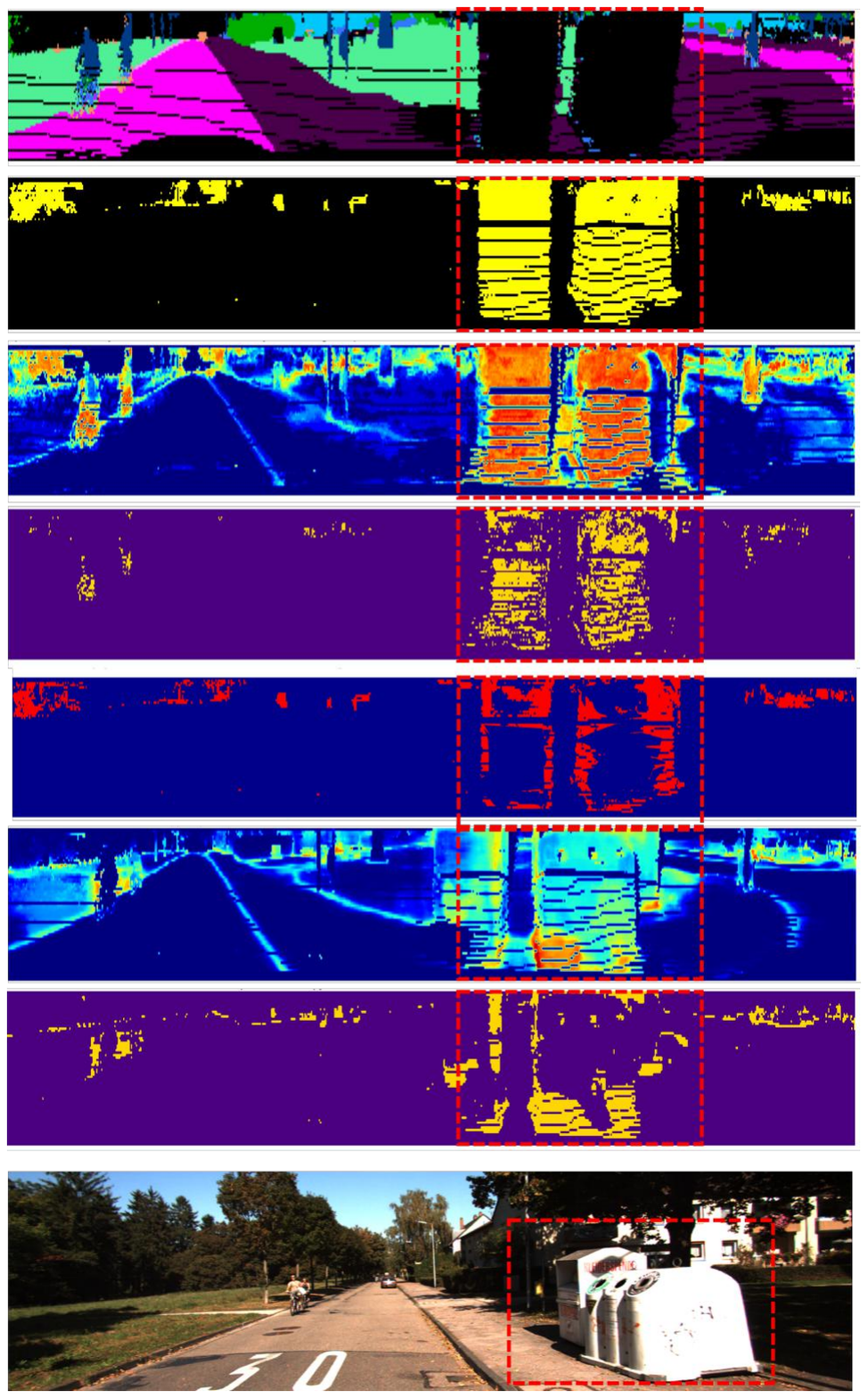}
        \caption{OOD object: Trash bins}
        \label{fig:OOD_comparison-b}
    \end{subfigure}

    \vspace{0.2cm}

    \begin{subfigure}{0.45\textwidth}
        \centering
        \includegraphics[width=\textwidth]{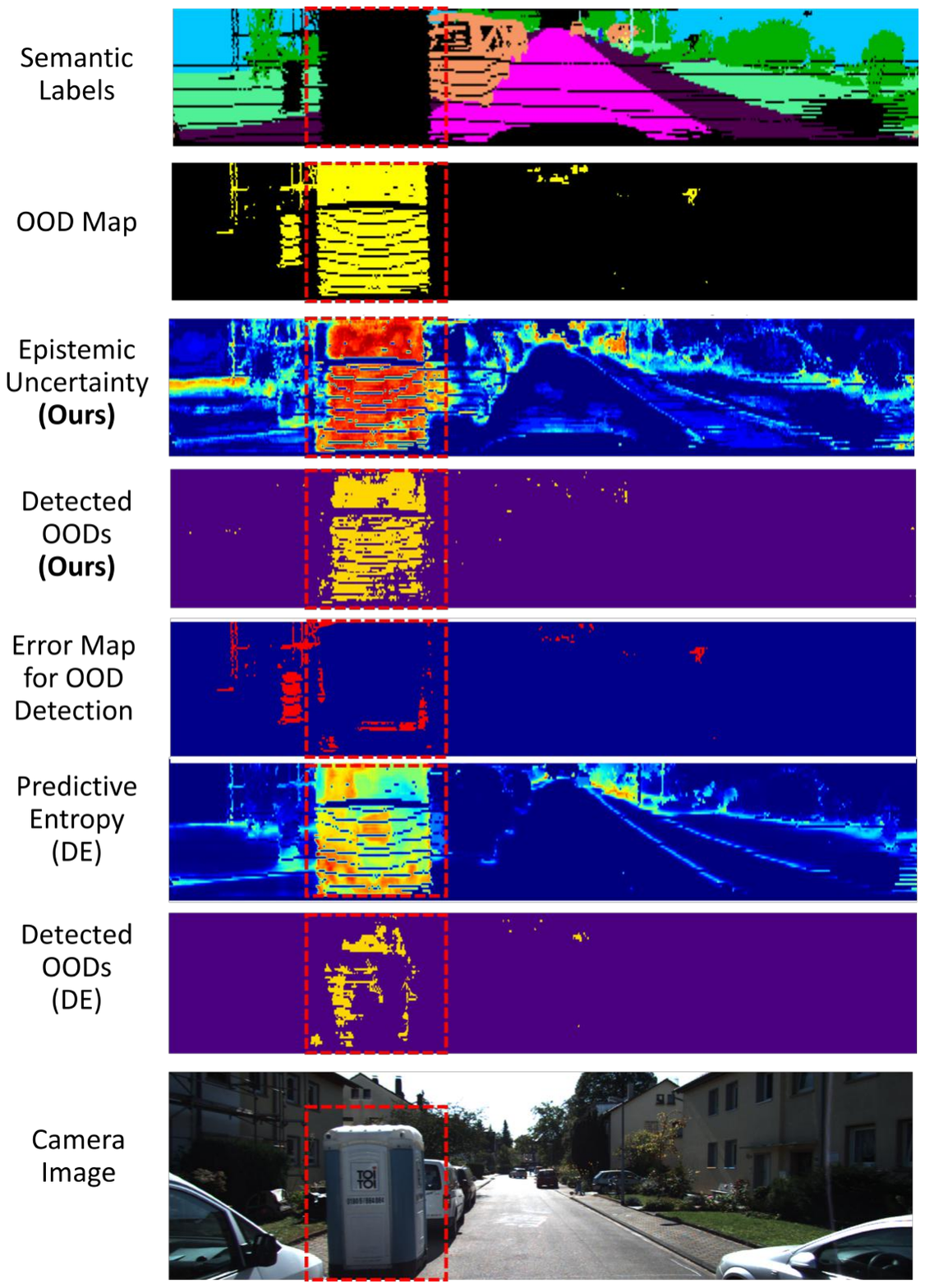}
        \caption{OOD object: Toilet cabin}
        \label{fig:OOD_comparison-c}
    \end{subfigure}
    \hfill
    \begin{subfigure}{0.45\textwidth}
        \centering
        \includegraphics[width=\textwidth]{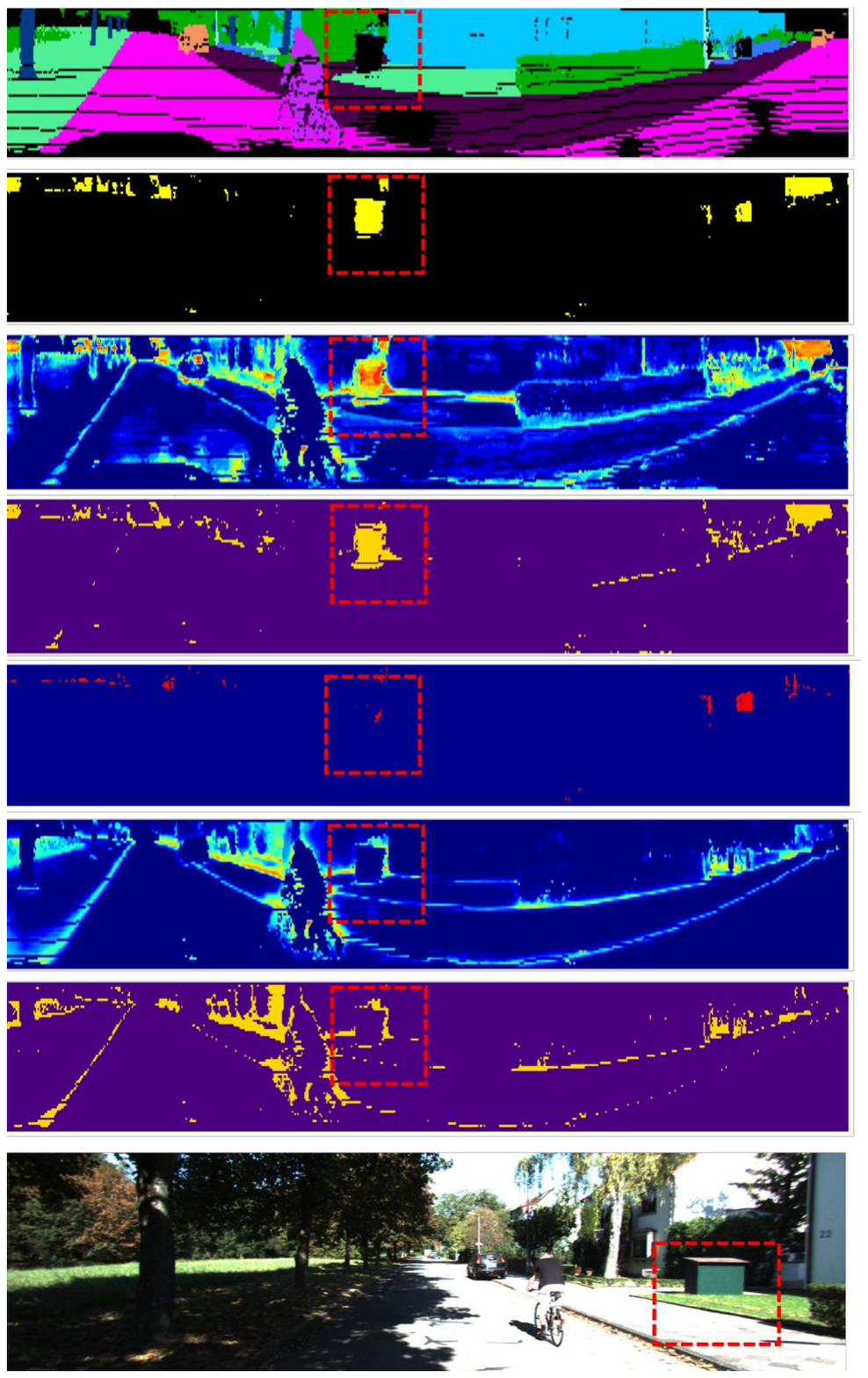}
        \caption{OOD object: Unknown structure}
        \label{fig:OOD_comparison-d}
    \end{subfigure}
    \caption{Qualitative comparison of OOD detection using epistemic uncertainty from our proposed approach versus predictive entropy from DE across four representative LiDAR scenes. From top to bottom, the rows depict: (1) Semantic predictions with class color coding: \colorbox{outlier}{\textcolor{white}{outlier}}, \colorbox{car}{\textcolor{white}{car}}, \colorbox{road}{\textcolor{white}{road}}, \colorbox{sidewalk}{\textcolor{white}{sidewalk}}, \colorbox{building}{\textcolor{white}{building}}, \colorbox{fence}{\textcolor{white}{fence}}, \colorbox{vegetation}{\textcolor{white}{vegetation}}, \colorbox{bicycle}{\textcolor{white}{trunk}}, \colorbox{terrain}{terrain}, \colorbox{pole}{pole}; (2) OOD ground truth, where ID pixels are black and OOD pixels are yellow; (3) Epistemic uncertainty map from our method and (5) predictive entropy map from DE, visualized with a temperature scale from \colorbox{blue}{\textcolor{white}{low}} to \colorbox{red}{\textcolor{white}{high}} uncertainty; (4) OOD detection results from our method and (7) DE, highlighting detected OOD regions in yellow; (5) Error map highlighting OOD objects not identified by our method, marked in red; (8) Camera image for visual reference. }
    \label{fig:OOD_comparison}
\end{figure}

Figure~\ref{fig:OOD_comparison} presents a qualitative comparison of OOD detection results between our proposed method (using epistemic uncertainty) and DE, across four representative LiDAR scenes. Each column corresponds to a different scene, while each row illustrates intermediate and final outputs, including semantic segmentation, OOD ground-truth (OODs are shown in yellow color), uncertainty maps, and detected OOD maps. The final row includes the corresponding RGB camera images for better interpretation of the scene content. OOD objects, such as placards, trash bins, and unknown structures, are highlighted with red dashed boxes across all images.

Overall, the detected OOD regions based on our epistemic uncertainty estimation (row 4) closely align with the ground-truth annotations (row 2). In particular, the epistemic uncertainty maps (row 3) show strong activations over truly unknown regions, while maintaining moderate uncertainty levels for misclassified ID samples. In contrast, predictive entropy maps from DE (row 5) suffer from over-activation due to their conflation of epistemic and aleatoric uncertainties. As a result, DE produces false positives in regions of high aleatoric uncertainty, especially along semantic boundaries, rather than reliably focusing on OOD objects.

In Figure~\ref{fig:OOD_comparison-a}, the placard is completely missed by OOD prediction from DE approach, while high predictive entropy arises along the sidewalk–vegetation border, leading to incorrect OOD detection. A similar misdetection occurs in Figure~\ref{fig:OOD_comparison-d}, where the boundary between the bicycle and street, as well as between terrain and road, is incorrectly identified as OOD. In these cases, the true OOD instances are either missed or partially detected.

In Figure~\ref{fig:OOD_comparison-a}, the placard is completely missed by OOD prediction from DE approach, while high predictive entropy arises along the sidewalk–vegetation border, leading to incorrect OOD detection. A similar misdetection occurs in Figure~\ref{fig:OOD_comparison-d}, where the boundary between the bicycle and street, as well as between terrain and road, is incorrectly identified as OOD. In these cases, the true OOD instances are either missed or partially detected. Figure~\ref{fig:OOD_comparison-b} also highlights the limitations of the DE approach in comparison to ours. While our method accurately detects the trash bins as OOD samples, DE only partially identifies them.

Figure~\ref{fig:OOD_comparison} reveals a minor shortcoming of our method: it assigns high epistemic uncertainty to misclassified ID objects in some cases. It reveals a few ID pixels misclassified as OOD by our approach, though these are significantly fewer than those misclassified by the DE method. This helps explain why, despite achieving the highest AUROC and AUPRC, our method's FPR@95 remains at 40.14\%, indicating that some ID pixels are still incorrectly flagged as OOD. This is primarily due to elevated epistemic uncertainty in semantically ambiguous or poorly segmented regions. For instance, thin structures, or mislabeled ID instances (e.g., fences, poles, or dynamic objects like bicyclists) tend to receive high epistemic uncertainty scores. These cases can lead to false positive detections, particularly when the uncertainty threshold is too sensitive to subtle representation variations.

\section{Conclusion}
In this work, we introduce an unsupervised OOD detection method in LiDAR semantic segmentation by employing the epistemic uncertainty, estimated from GMMs in the feature space of a deep classifier. Our approach addresses key limitations of existing predictive entropy-based approaches, which often misidentify ambiguous ID samples as OOD. By employing epistemic uncertainty, we demonstrate improved discrimination between ID and OOD data. Experimental results on SemanticKITTI validate that our approach surpasses existing methods such as MSP, ODIN, MC Dropout, and deep ensembles, achieving the highest AUROC, highest AUPRC, lowest FPR95, and improved segmentation accuracy (mIoU). Despite these advancements, our approach still exhibits some false positives due to elevated uncertainty in complex boundary regions and misclassified objects. Future work will investigate strategies to reduce such ambiguity and refine the uncertainty threshold selection to further enhance real-world applicability.


%
%
%
%
\bibliographystyle{splncs04}
\bibliography{073-main.bib}

\begin{thebibliography}{10}
\providecommand{\url}[1]{\texttt{#1}}
\providecommand{\urlprefix}{URL }
\providecommand{\doi}[1]{https://doi.org/#1}

\bibitem{behley2019semantickitti}
Behley, J., Garbade, M., Milioto, A., Quenzel, J., Behnke, S., Stachniss, C.,
  Gall, J.: Semantickitti: A dataset for semantic scene understanding of lidar
  sequences. In: Proceedings of the IEEE/CVF international conference on
  computer vision. pp. 9297--9307 (2019)

\bibitem{bevandic2018discriminative}
Bevandi{\'c}, P., Kre{\v{s}}o, I., Or{\v{s}}i{\'c}, M., {\v{S}}egvi{\'c}, S.:
  Discriminative out-of-distribution detection for semantic segmentation. arXiv
  preprint arXiv:1808.07703  (2018)

\bibitem{bevandic2019simultaneous}
Bevandi{\'c}, P., Kre{\v{s}}o, I., Or{\v{s}}i{\'c}, M., {\v{S}}egvi{\'c}, S.:
  Simultaneous semantic segmentation and outlier detection in presence of
  domain shift. In: Pattern Recognition: 41st DAGM German Conference, DAGM GCPR
  2019, Dortmund, Germany, September 10--13, 2019, Proceedings 41. pp. 33--47.
  Springer (2019)

\bibitem{chawla2004special}
Chawla, N.V., Japkowicz, N., Kotcz, A.: Special issue on learning from
  imbalanced data sets. ACM SIGKDD explorations newsletter  \textbf{6}(1),
  ~1--6 (2004)

\bibitem{cortinhal2020salsanext}
Cortinhal, T., Tzelepis, G., Erdal~Aksoy, E.: Salsanext: Fast,
  uncertainty-aware semantic segmentation of lidar point clouds. In: Advances
  in Visual Computing: 15th International Symposium, ISVC 2020, San Diego, CA,
  USA, October 5--7, 2020, Proceedings, Part II 15. pp. 207--222. Springer
  (2020)

\bibitem{di2021pixel}
Di~Biase, G., Blum, H., Siegwart, R., Cadena, C.: Pixel-wise anomaly detection
  in complex driving scenes. In: Proceedings of the IEEE/CVF conference on
  computer vision and pattern recognition. pp. 16918--16927 (2021)

\bibitem{gal2015dropout}
Gal, Y., Ghahramani, Z.: Dropout as a bayesian approximation. arXiv preprint
  arXiv:1506.02157  (2015)

\bibitem{guo2017calibration}
Guo, C., Pleiss, G., Sun, Y., Weinberger, K.Q.: On calibration of modern neural
  networks. In: International conference on machine learning. pp. 1321--1330.
  PMLR (2017)

\bibitem{hendrycks2016baseline}
Hendrycks, D., Gimpel, K.: A baseline for detecting misclassified and
  out-of-distribution examples in neural networks. arXiv preprint
  arXiv:1610.02136  (2016)

\bibitem{hendrycks2018deep}
Hendrycks, D., Mazeika, M., Dietterich, T.: Deep anomaly detection with outlier
  exposure. arXiv preprint arXiv:1812.04606  (2018)

\bibitem{hendrycks2019using}
Hendrycks, D., Mazeika, M., Kadavath, S., Song, D.: Using self-supervised
  learning can improve model robustness and uncertainty. Advances in neural
  information processing systems  \textbf{32} (2019)

\bibitem{hornauer2023heatmap}
Hornauer, J., Belagiannis, V.: Heatmap-based out-of-distribution detection. In:
  Proceedings of the IEEE/CVF Winter Conference on Applications of Computer
  Vision. pp. 2603--2612 (2023)

\bibitem{huang2022out}
Huang, C., Abdelzad, V., Mannes, C.G., Rowe, L., Therien, B., Salay, R.,
  Czarnecki, K., et~al.: Out-of-distribution detection for lidar-based 3d
  object detection. In: 2022 IEEE 25th International Conference on Intelligent
  Transportation Systems (ITSC). pp. 4265--4271. IEEE (2022)

\bibitem{jiang2018trust}
Jiang, H., Kim, B., Guan, M., Gupta, M.: To trust or not to trust a classifier.
  Advances in neural information processing systems  \textbf{31} (2018)

\bibitem{kirsch2021pitfalls}
Kirsch, A., Mukhoti, J., van Amersfoort, J., Torr, P.H., Gal, Y.: On pitfalls
  in ood detection: entropy considered harmful. In: Uncertainty \& Robustness
  in Deep Learning Workshop. ICML (2021)

\bibitem{lakshminarayanan2017simple}
Lakshminarayanan, B., Pritzel, A., Blundell, C.: Simple and scalable predictive
  uncertainty estimation using deep ensembles. Advances in neural information
  processing systems  \textbf{30} (2017)

\bibitem{lee2018simple}
Lee, K., Lee, K., Lee, H., Shin, J.: A simple unified framework for detecting
  out-of-distribution samples and adversarial attacks. Advances in neural
  information processing systems  \textbf{31} (2018)

\bibitem{liang2022gmmseg}
Liang, C., Wang, W., Miao, J., Yang, Y.: Gmmseg: Gaussian mixture based
  generative semantic segmentation models. Advances in Neural Information
  Processing Systems  \textbf{35},  31360--31375 (2022)

\bibitem{liang2017enhancing}
Liang, S., Li, Y., Srikant, R.: Enhancing the reliability of
  out-of-distribution image detection in neural networks. arXiv preprint
  arXiv:1706.02690  (2017)

\bibitem{minderer2021revisiting}
Minderer, M., Djolonga, J., Romijnders, R., Hubis, F., Zhai, X., Houlsby, N.,
  Tran, D., Lucic, M.: Revisiting the calibration of modern neural networks.
  Advances in neural information processing systems  \textbf{34},  15682--15694
  (2021)

\bibitem{nguyen2015deep}
Nguyen, A., Yosinski, J., Clune, J.: Deep neural networks are easily fooled:
  High confidence predictions for unrecognizable images. In: Proceedings of the
  IEEE conference on computer vision and pattern recognition. pp. 427--436
  (2015)

\bibitem{rezende2015variational}
Rezende, D., Mohamed, S.: Variational inference with normalizing flows. In:
  International conference on machine learning. pp. 1530--1538. PMLR (2015)

\bibitem{shojaei2025hierarchical}
Shojaei, H.: Hierarchical bayesian modeling of epistemic uncertainty in lidar
  semantic segmentation (Jun 2025). \doi{10.5281/zenodo.15635087},
  \url{https://doi.org/10.5281/zenodo.15635087}

\bibitem{shojaei2024uncertainty}
Shojaei~Miandashti, H., Zou, Q., Mehltretter, M.: Uncertainty estimation and
  out-of-distribution detection for lidar scene semantic segmentation. In:
  European Conference on Computer Vision. pp. 116--131. Springer (2024)

\bibitem{smith2018understanding}
Smith, L., Gal, Y.: Understanding measures of uncertainty for adversarial
  example detection. arXiv preprint arXiv:1803.08533  (2018)

\bibitem{vandenhende2021multi}
Vandenhende, S., Georgoulis, S., Van~Gansbeke, W., Proesmans, M., Dai, D.,
  Van~Gool, L.: Multi-task learning for dense prediction tasks: A survey. IEEE
  transactions on pattern analysis and machine intelligence  \textbf{44}(7),
  3614--3633 (2021)

\bibitem{vojivr2024pixood}
Voj{\'\i}{\v{r}}, T., {\v{S}}ochman, J., Matas, J.: Pixood: Pixel-level
  out-of-distribution detection. In: European Conference on Computer Vision.
  pp. 93--109. Springer (2024)

\bibitem{williams2021fool}
Williams, D.S., Gadd, M., De~Martini, D., Newman, P.: Fool me once: Robust
  selective segmentation via out-of-distribution detection with contrastive
  learning. In: 2021 IEEE International Conference on Robotics and Automation
  (ICRA). pp. 9536--9542. IEEE (2021)

\end{thebibliography}

\end{document}